\pdfoutput=1

\documentclass[11pt]{article}

\usepackage[final]{ACL2023}

\usepackage{times}
\usepackage{latexsym}

\usepackage[T1]{fontenc}

\usepackage[utf8]{inputenc}

\usepackage{microtype}

\usepackage{inconsolata}

\usepackage{graphicx}
\usepackage{subcaption}
\usepackage{booktabs}
\usepackage{graphicx}
\usepackage{tikz}
\usepackage{amsmath} 
\usepackage{booktabs}
\usepackage{multirow}
\usepackage{makecell}
\usepackage{caption}
\usepackage{amssymb}
\usepackage{lipsum,afterpage,refcount}
\usepackage{graphicx}              
\usepackage{amsmath,amssymb}
\usepackage{tikz}
\usepackage{pgfplots}
\pgfplotsset{compat=1.18}

\usepackage{svg}
\title{p\textsuperscript{2}-TQA: A Process-based Preference Learning Framework \\ for Self-Improving Table Question Answering Models}

\author{Wei Zhou$^{1,3}$ \hspace{5mm}
  Mohsen Mesgar$^1$ \hspace{5mm}
  Heike Adel$^{2}$\hspace{5mm} 
  Annemarie Friedrich$^3$
    \\
  $^1$Bosch Center for Artificial Intelligence, Renningen, Germany \\ 
      $^2$Hochschule der Medien, Stuttgart, Germany\hspace{2.0mm} $^3$University of Augsburg, Germany \hspace{5mm} \\
\texttt{\{wei.zhou3|mohsen.mesgar\}@de.bosch.com}\\ 
  \texttt{annemarie.friedrich@uni-a.de} \hspace{5mm} \texttt{adel-vu@hdm-stuttgart.de}}

\begin{document}
\maketitle
\begin{abstract}


Table question answering (TQA) focuses on answering questions based on tabular data. 
Developing TQA systems targets effective interaction with tabular data for tasks such as cell retrieval and data analysis. 
While recent work has leveraged fine-tuning to improve TQA systems, existing approaches often under-utilize available data and neglect the potential of post-training for further gains.
In this work, we introduce p\textsuperscript{2}-TQA,  a \underline{p}rocess-based \underline{p}reference learning framework for TQA post-training. 
p\textsuperscript{2}-TQA automatically constructs process-based preference data via a table-specific pipeline, eliminating the need for manual or costly data collection. 
It then optimizes models through contrastive learning on the collected data.
Experiments show that p\textsuperscript{2}-TQA effectively improves TQA models by up to 5\% on in-domain datasets and 2.4\% on out-of-domain datasets with only 8,000 training instances.
Furthermore, models enhanced with p\textsuperscript{2}-TQA achieve competitive results against larger, more complex state-of-the-art TQA systems, while maintaining up to five times higher efficiency.
\end{abstract}

\section{Introduction}
Table question answering (TQA) aims to generate accurate responses to queries over tables.  
Current TQA systems fall into two categories: fine-tuned models \cite{zhang2025tablellmenablingtabulardata, wu-feng-2024-protrix} and training-free frameworks \cite{zhou-etal-2025-efficient, nahid-rafiei-2024-tabsqlify}. 
The former fine-tune pre-trained small-size ($\leq$ 8B) large language models (LLMs)
while the latter rely on large LLMs and involve complex designs.
There is a growing interest in understanding and developing fine-tuned TQA models \cite{deng2025betterunderstandingtableinstruction,deng2025rethinkingtableinstructiontuning} due to their promising performance and inference efficiency.

\begin{figure}
    \centering
    \includegraphics[width=1\linewidth]{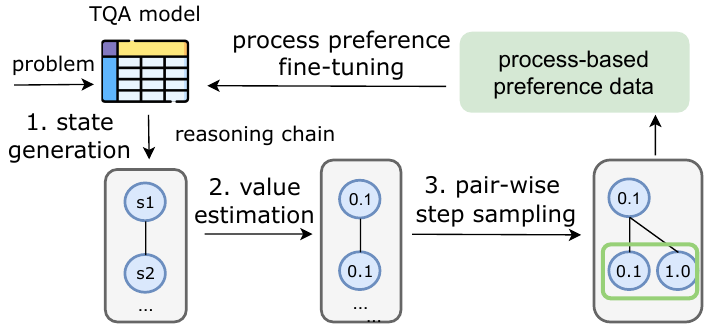}
    \caption{An overview of p\textsuperscript{2}-TQA: An existing model generates reasoning chains for a given problem. The chains are parsed into states, composed of cumulative steps. Each state is scored by a value function. We then create pairwise steps by rolling out parent states, selecting those with value differences exceeding a threshold. Lastly, contrastive learning is performed over collected data to improve the TQA model.}
    \label{fig:overview}
\end{figure}

Current fine-tuning methods for TQA often augment only a subset of existing datasets with \text{Chain-of-Thought (CoT)} reasoning \cite{wei2023chainofthoughtpromptingelicitsreasoning}, constrained by the high cost of querying large commercial models \cite{wu-feng-2024-protrix}. Models are then trained on these reasoning chains and answers via supervised fine-tuning. 
This workflow has two key limitations:
(1) only part of the training data is used, i.e., data is under-utilized; and
(2) potential performance gains from post-training are neglected.
Based on the identified gaps, this study aims to answer the research question: \textit{How can we leverage existing datasets to post-train a TQA model for further performance gains?}
Our effective post-training method (shown in Figure~\ref{fig:overview}) enhances models without requiring additional manual or costly data.


While post-training with self-generated data has primarily been explored in mathematics and coding \cite{singh2024humandatascalingselftraining, zelikman2024selftaughtoptimizerstoprecursively,xiong-etal-2024-watch, he2024semisupervisedrewardmodelingiterative}, 
particularly through step-wise preference learning \cite{tu2025enhancingllmreasoningiterative,xu-etal-2024-chatglm}, its application in TQA remains under-explored.
Previous methods for obtaining step-wise preference pairs typically rely on (1) closed-source or large open-source models as judges to discern correct and incorrect steps \cite{lai2024stepdpostepwisepreferenceoptimization}, (2) Monte Carlo Sampling (MCS) to estimate a step's quality  \cite{wang-etal-2024-math, xiong-etal-2024-watch, hwang-etal-2024-self}, or (3) a combination of both  \cite{zhang2025lessonsdevelopingprocessreward}. 
However, performing step-wise preference learning to TQA presents unique challenges.
Compared to mathematics, it typically involves longer inputs and intermediate reasoning due to large tables.
This necessitates a careful design of steps; simply using newline breaks to obtain steps, as in math, could lead to excessive computation costs, because newline breaks also denote new rows in tables.
Furthermore, the structured nature of input in TQA demands a reconsideration of value functions, since LLM judges have a limited understanding of table structures \cite{sui2024tablemeetsllmlarge}, especially with larger tables \cite{zhou-etal-2024-freb}.

To this end, we introduce p\textsuperscript{2}-TQA, a \underline{p}rocess-based \underline{p}reference learning pipeline for TQA post-training:  
p\textsuperscript{2}-TQA operates in three stages for data collection (Figure \ref{fig:overview}): \textit{state generation}, \textit{state value estimation}, and \textit{pair-wise step sampling}.
The first stage collects and parses reasoning chains into states, composed of cumulative reasoning steps that are carefully designed for TQA. 
The last two stages construct step-wise preference pairs via MCS for state value estimation and a stringent filtering process for quality control. 
We apply direct preference optimization (DPO) \cite{rafailov2024directpreferenceoptimizationlanguage} with the collected data to self-improve a TQA model.

Experiments show that p\textsuperscript{2}-TQA improves TQA models by up to 5\% on in-domain datasets and by up to 2.4\% on out-of-domain datasets with only 8k preference training pairs.
It surpasses methods that require additional LLMs as judges, yet it is ten times more efficient. 
This underscores our contribution in establishing an effective and efficient framework for self-improving TQA models.
The self-improved models outperform existing fine-tuned TQA models and achieve comparable performance to much larger and more complex frameworks on three datasets, while maintaining five times higher inference efficiency.
Code is available. \footnote{\url{https://github.com/boschresearch/p2-TQA}}

\section{Step-wise Preference Learning for TQA}
Given a table $t$, a question $q$, and a fine-tuned TQA model $M_{ft}$ that outputs a reasoning chain $r$ consisting of $l$ steps: $\{k_1,k_2,...,k_l\}$, along with a predicted answer $a$,  our goal is to collect high-quality step-wise preference data $(k_i^{good}, k_i^{bad})$ from $M_{ft}$ and perform contrastive learning to improve $M_{ft}$.

\paragraph{Step Design and State Generation.}
Designing an effective step scope is crucial for both performance and sampling efficiency.
Inspired by SQL query operations, we define a step as a basic operation, such as filtering or counting. 
To aid LLM reasoning in TQA, each step includes a planning component that outlines the required operation and a reasoning component that provides its result, e.g., \textit{Count the number of gold medals received in 2004. There are 6 gold medals received in 2004.}
For each TQA problem, an initial state $s_0$ is formed from $t$, $q$, and an instruction $u$.
We then sample $m$ reasoning chains, denoted as $\{r\}_{i=1}^m$ from $M_{ft}$. 
Prompts can be found in Appendix \ref{prompts}. 
A new state $s_i$ is defined as the combination of previous state $s_{i-1}$ and a step $k_i$ generated at the timestep $i$: $s_i=(s_{i-1}, k_i)$, where $s_{i-1}=\{s_0, k_1,k_2,....k_{i-1}\}$.
Problems where all reasoning traces lead to correct answers are discarded, as they are considered too easy.

\begin{figure}
    \centering
    \includegraphics[width=0.9\linewidth]{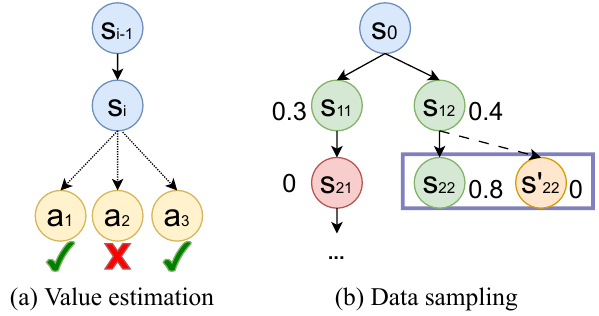}
    \caption{Process-based preference data collection. We estimate a state value by the probability of a state leading to a correct answer. In the first example, $V(s_i)=\frac{2}{3}$. After obtaining state values, we do not consider intermediate states that have a value of 0 ($s_{21}$), together with their child states. We sample pair-wise states for each remaining state, e.g., $s'_{22}$ is sampled by rolling out $s_{12}$ and is regarded as a pair state for $s_{22}$.}
    \label{fig:sample}
\end{figure}

\paragraph{State Value Estimation.}
\label{sec:svf}
A state value function $V$ takes in a state and returns its value. 
We approximate a state's value by using Monte Carlo Sampling similar to \citet{wang-etal-2024-math}: $M_{ft}$ takes in $s_i$ and completes the current reasoning chain until reaching an answer. This is repeated $n$ times.
$V(s_i)$ is calculated as the probability of $s_i$ leading to the correct answer.
An example is shown in Figure \ref{fig:sample}, where $V(s_i)=\frac{2}{3}$.
The continuous value allows a flexible and controlled selection of pair-wise steps described in the following paragraph.

 
\paragraph{Pair-wise Step Sampling.}

After obtaining state values, we filter out intermediate states $s_i$, where $V(s_{i}) =0$, and also remove their child states $\{s_{i+1},...s_{z}\}$.
This is exemplified by the red nodes in Figure~\ref{fig:sample}. 
We assume a state of value 0 to contain erroneous steps. Rolling out from it is likely to create bad-quality child states.
For each remaining state $s_i$, we use the completion traces sampled when calculating $V(s_{i-1})$: $\{(s_{i,j},...,s_{z_{j},j}, a_j)\}_{j=1}^n$ as rollouts, where $a_j$ is the predicted answer and $z_j$ is the total number of steps for the $j$-th finalized solution.
Next, we calculate state values for each sampled $\{s_{i,j}\}_{j=1}^n$. This results in a set of pair-wise states: $(s_i, \{s_{i,j}\}_{j=1}^n)$ that can be used to construct step-wise preference dataset $D_{sdpo}$.
As $V(s_i)$ is a continuous value, a pair comprising one good state $s_{good}$ and one bad state $s_{bad}$ is selected if $V(s_{good})-V(s_{bad})\geq \tau$, where $\tau$ is a hyper-parameter.
We prove later in our experiments  that for TQA, this filtering mechanism greatly improves performance and efficiency on top of using MCS as the value function, while maintaining efficiency.
The preference data for step DPO can be represented as $D_{sdpo} = \{( s_{i-1}, k_i^{good}, k_i^{bad})^{d}\}_{d=1}^{|D_{sdpo}|}$. 

After collecting the preference dataset, we fine-tune $M_{ft}$ using pairs of good and bad steps given previous steps. 
The loss function is defined as follows, where $\beta$ is a hyper-parameter controlling the strength of incorporating the preference signal.
$\pi_{ref}$ and $\pi_\theta$ denote the original reference and updated model, respectively.
\begin{equation}
\scalebox{0.8}{
\begin{math}
\begin{aligned}
    \mathcal{L} = -\mathbb{E}_{(s_{i-1}, k_i^{good}, k_i^{bad}) \sim \mathcal{D}_{sdpo}} \left[ \log \sigma \left( \beta \log \frac{\pi_\theta(k_i^{good} | s_{i-1})}{\pi_{ref}(k_i^{good} | s_{i-1})} \right) \right. \\
\left. - \beta \log \frac{\pi_\theta(k_i^{bad} | s_{i-1})}{\pi_{ref}(k_i^{bad} | s_{i-1})} \right]
 \end{aligned}
 \end{math}
}
\end{equation}


\section{Experiments}
We present details for TQA models, baselines, datasets, and experimental settings in this section.

\paragraph{TQA Models.} Existing fine-tuned TQA models do not feature clear step separations. We therefore obtain a TQA model $M_{ft}$ by fine-tuning an LLM using the step definition introduced before. 
Following previous work \cite{wu-feng-2024-protrix, zhang2025tablellmenablingtabulardata}, we employ Deepseek-V3 \cite{deepseekai2025deepseekv3technicalreport} to generate reasoning chains. 
We sample 2.4k, 1.5k, and 2.3k examples from the training sets of WTQ \cite{pasupat-liang-2015-compositional}, TabFact \cite{2019TabFactA}, and HiTab \cite{cheng-etal-2022-hitab}, respectively. 
We prompt Deepseek-V3 to produce reasoning chains along with final answers, retaining only those chains that yield correct answers. 
This process results in 1,612 instances from WTQ, 1,425 from TabFact, and 1,277 from HiTab, for a total of 4,314 instances.

\paragraph{Baselines.}
We consider the following baselines for \textbf{self-improvement strategies}: 
(1) \textit{RFT} \cite{yuan2023scalingrelationshiplearningmathematical} trains a model with self-generated reasoning traces that lead to correct answers using supervised fine-tuning. 
(2) \textit{FDPO} \cite{xu-etal-2024-chatglm} trains a model with pair-wise correct and incorrect full reasoning chains using DPO. 

Baselines for \textbf{value functions} include: 
(3) \textit{MC with binary labels} (\textsc{mc-b}) \cite{wang-etal-2024-math} returns binary state values based on whether states derive final correct answers. 
\textit{Mixed estimation} (\textsc{mix}) \cite{zhang2025lessonsdevelopingprocessreward} scores $s_i$ 1 if both \textsc{mc-b} and an external LLM judge $M_j$ output 1. 
If both judges return 0, $s_i$ is 0.
States receiving different scores from the judges are not considered for building the preference dataset. 
(4) \textsc{Self-Explore} \cite{hwang-etal-2024-self} randomly selects a preferred reasoning trace and uses full completion of it instead of a step. This results in longer preferred responses over rejected ones.  

Baselines for \textbf{TQA models} include both end-to-end and training-free frameworks: (1) TableLlaMA \cite{Zhang2023TableLlamaTO} is an end-to-end fine-tuned model with LlaMA-2-7B \cite{Touvron2023Llama2O} as the base model.
(2) Protrix \cite{wu-feng-2024-protrix} is fine-tuned with around 4k instances with reasoning chains generated from GPT-4, also using LlaMA-2-7B as the base model.
(3) MACT \cite{zhou-etal-2025-efficient} is a training-free framework, leveraging tools and agent collaboration.
(4) TabSQLify \cite{nahid-rafiei-2024-tabsqlify} decomposes tables into relevant sub-tables with SQL query generation and execution. Then, sub-tables and questions are passed to LLMs to obtain final answers.

\paragraph{Datasets.}
We train $M_{ft}$ using the training sets of WTQ, TabFact, and HiTab. To obtain preference data, we sample from their validation sets. We use the test sets of these three datasets as in-domain evaluation data and incorporate three out-of-domain datasets:  WikiSQL \citep{zhongSeq2SQL2017}, SCITAB \cite{lu-etal-2023-scitab}, and CRT \cite{zhang-etal-2023-crt} to test models' generalisability.
These datasets test our method in various degrees of complexity. Thus, we make sure our method is generally effective.
Details about the datasets are presented in Appendix \ref{datasets}.


\paragraph{Experimental Settings.}
\label{sec:experimental_settings}
We choose Qwen-2.5-7B \cite{qwen2025qwen25technicalreport} and LlaMA-3.1-8B \cite{grattafiori2024llama3herdmodels} as base models.
During preference learning, we fix the fine-tuning dataset size to 8k for Qwen-2.5-7B and 6.7k for LlaMA-3.1-8B, as different baselines result in different sample sizes (statistics are shown in Appendix \ref{sampled_data}). 
For fair comparison, we use the smallest sample size collected as the fine-tuning data size. 
Hyper-parameters are shown in Appendix \ref{hyperparameters}. We use Qwen-2.5-72B as $M_j$ (Prompt in Appendix \ref{prompts}). 
The number of reasoning chains $m$ is set to 4, and the roll-out number $n$ is set to 8. 
The threshold $\tau$ is set to 0.9, and the temperature is set to 0.7 and 0 during dataset construction and inference, respectively.
We use Exact Match (EM) as the evaluation metric. All experiments are conducted using 4 A100 GPUs.   Training is performed with LlaMA-Factory \cite{zheng2024llamafactory} and inference is performed with VLLM \cite{kwon2023efficient}.

\section{Results and Discussions}
\begin{figure}
    \centering
    \includegraphics[width=1.03\linewidth]{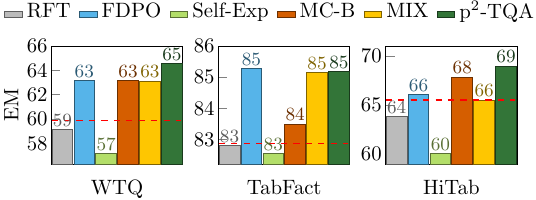}
    \caption{Comparing  p\textsuperscript{2}-TQA with baselines using Exact Match. Results are averaged across models. RFT and FDPO stand for rejected sampling fine-tuning and full-chain DPO, respectively. We experiment with several value functions: \textsc{Self-Exp} (Self-Exploration), \textsc{mc-b} (Monte Carlo with binary values), and \textsc{mix} (a combination of LLM-as-a-judge and \textsc{mc-b}). Dashed lines show performances of fine-tuned TQA models $M_{ft}$ before applying self-improvement methods.}
    \label{fig:results}
\end{figure}

Figure \ref{fig:results} shows the Exact Match of different methods on in-domain datasets, averaged across models. Per-model results can be found in Appendix \ref{additional_results}. 

\paragraph{p\textsuperscript{2}-TQA effectively improves the performance of TQA models.} As Figure \ref{fig:results} shows, applying \text{p\textsuperscript{2}-TQA} enhances the performance of $M_{ft}$
by 3.5\% on average on in-domain datasets.
The gains are more obvious on WTQ (5\%) compared to TabFact (2\%). This might be attributed to dataset features: TabFact is a binary classification dataset, thus it is easier for models to achieve high performance and harder to further reach improvements.  
When evaluating on out-of-domain datasets, we witness an average of 2.2\% performance gain after applying our framework ({\textcolor{black}{$\uparrow$}} 2.2\% for WikiSQL, {\textcolor{black}{$\uparrow$}} 2.4\% for SCITAB, and {\textcolor{black}{$\uparrow$}} 2.1\% for CRT).
These findings demonstrate the generalisability of our method on out-of-domain data. 
The performance gains on both in-domain and out-of-domain datasets are significant and can be observed with inference sampling over multiple runs. Detailed results are shown in Appendix \ref{additional_results}.

\begin{table}[t]
    \small
    \setlength{\tabcolsep}{1.2pt}
    \centering    
 
    \begin{tabular}{lcccccc}
    \toprule
    Models & WTQ & TabFact &HiTab & WikiSQL & SCITAB & CRT\\
    \midrule
 Protrix & 56.2 & 71.6&- &67.4  &45.0&40.2\\
 T-LlaMA & 35.0&82.6&64.7 & 50.5&38.6&26.9\\
  $M_{si}$-Qwen & 63.1 & \underline{84.9} & \underline{67.6}&\underline{72.0} &\textbf{56.9}&\underline{51.4}\\
  $M_{si}$-LlaMA & \underline{65.8} & \textbf{85.5} & \textbf{70.3}&70.8&54.4&50.3\\
 \midrule
 
 MACT  & \textbf{70.4} & - &- & -& \underline{55.8}& \textbf{57.4} \\
 T-SQLify & 64.7 & 80.2 &- &\textbf{76.7} &50.9 & 42.0 \\
    \bottomrule
    \end{tabular}
       \caption{Exact Match of TQA models. $M_{si}$ refers to self-improved models. 
       T-LlaMA and T-SQLify refer to TableLlaMA and TabSQLify, respectively.
       Framework results (the last two rows) are obtained using GPT-3.5 as the backbone. State-of-the-art TQA results are obtained from previous work \cite{zhou-etal-2024-freb, Zhang2023TableLlamaTO, wu-feng-2024-protrix, nahid-rafiei-2024-tabsqlify}.\label{tab:results_tqa}}
\end{table}

\paragraph{p\textsuperscript{2}-TQA delivers competitive results against baselines, highlighting the effectiveness of pairing a lightweight value function with stringent filtering for self-improving models.} 
Comparing p\textsuperscript{2}-TQA with RFT and FDPO, we find that our framework leads to higher improvements. 
The effect is more obvious on in-domain datasets than out-of-domain datasets, as results in Appendix \ref{additional_results} show. 
Though FDPO is generally computationally cheaper than p\textsuperscript{2}-TQA, i.e., under the same token budgets, it generates more training instances. We find that the performance of p\textsuperscript{2}-TQA improves when training example sizes increase. 
In contrast, more examples do not necessarily lead to better performance using FDPO, suggesting early saturation. 
More analysis is presented in Appendix \ref{cost_effectiveness}.

When comparing against methods using different value functions, we observe that p\textsuperscript{2}-TQA greatly outperforms \textsc{Self-Explore}.
It also shows advantages over methods that simply use a binary value function (\textsc{mc-b}), or combining it with an LLM judge (\textsc{mix}).
Notably, p\textsuperscript{2}-TQA takes 10 times less time than using \textsc{mix} when sampling the same amount of data.
This demonstrates that our method delivers strong performance while being efficient. 
Interestingly, the efficiency is not compromised for reasoning correctness. 
We evaluate models fine-tuned using data generated by p\textsuperscript{2}-TQA and \textsc{mix} with regard to step correctness and find that the two methods achieve similar accuracy (95.7\% vs.\ 94.6\%). 
Detailed analysis can be found in Appendix \ref{judge_analysis}. 
We provide an analysis of threshold impact in Appendix \ref{thresholds_analysis}, showing the necessity of picking a relatively high threshold for data filtering.

\paragraph{Self-improved TQA models achieve competitive results compared to complicated state-of-the-art approaches, with five times less inference time.}
Table \ref{tab:results_tqa} shows results for current TQA models. 
first four rows of Table \ref{tab:results_tqa} compare small-size fine-tuned TQA models, and the last two rows show state-of-the-art training-free frameworks back-boned by GPT-3.5. 
We find that both the Qwen and LlaMA models, enhanced using  \text{p\textsuperscript{2}-TQA}, outperform existing TQA models.
More importantly, both self-improved models achieve competitive performance compared to larger and more complex frameworks where tools and agentic collaboration are involved. 
On SCITAB, $M_{si}$ with the Qwen even achieves the best performance. 
Apart from the competitive task performance, we emphasize the inference efficiency of self-improved models: they require eight times less inference time than MACT and five times less than TabSQLify.

\begin{table}[t]
    \footnotesize
    \setlength{\tabcolsep}{3pt}
    \centering    
 
    \begin{tabular}{lccc}
    \toprule
    \textbf{Model} & \textbf{Retrieval} & \textbf{Reasoning} & \textbf{Total} \\
    \# instances & 1133 & 451 & 1584 \\ 
    \midrule
    Qwen ($M_{ft}$)  & 71.32 &36.36 & 61.36  \\
    Qwen ($M_{ft}$) +p\textsuperscript{2}-TQA  & \textbf{78.02} & \textbf{41.24} & \textbf{67.55}\\ \midrule
    LlaMA ($M_{ft}$) & 79.44 & \textbf{45.23}& 69.70 \\
    LlaMA ($M_{ft}$) +p\textsuperscript{2}-TQA& \textbf{81.38} & 42.57 & \textbf{70.32} \\

    \bottomrule
    \end{tabular}
       \caption{Exact Match of models with and without p\textsuperscript{2}-TQA,  evaluated across different question types on the HiTab test set. }
       \label{tab:error_analysis}
\end{table}

\begin{figure}
    \centering
    \includegraphics[width=0.85\linewidth]{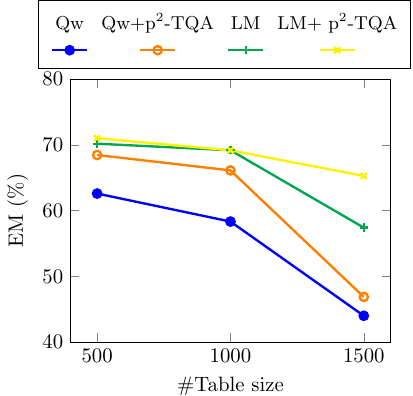}
    \caption{Exact Match of models with and without p\textsuperscript{2}-TQA,
    evaluated across different table sizes and averaged over in-domain datasets. Qw and LM show the performance of Qwen $M{ft}$ and LlaMA $M{ft}$ respectively.
    Instances are grouped into three bins by table token count: $<500$, $500$--$1000$, and $\ge 1000$.  }
    \label{fig:error_analysis}
\end{figure}

\paragraph{Applying p\textsuperscript{2}-TQA generally improves accuracy across question types, table sizes, and step correctness.}
We compare models fine-tuned with p\textsuperscript{2}-TQA against those without, from two perspectives: We take a closer look at question type and table size, inspired by \citet{zhou-etal-2025-texts}. 
For question type analysis, we categorize questions into those requiring only retrieval and those requiring reasoning in addition. We evaluate models on the HiTab dataset, which provides explicit question type annotations. 
As shown in Table \ref{tab:error_analysis}, the enhanced Qwen-2.5-7B model outperforms its pre-trained counterpart in both categories, whereas the LLaMA-3.1-8B model shows notable gains primarily in retrieval questions.
For table size analysis, we partition tables into three bins based on token count and compute EM accuracy for each bin.
Figure \ref{fig:error_analysis} reports results averaged over in-domain datasets, revealing that self-improved models consistently achieve higher accuracy across all table sizes.

Finally, we examine step correctness by comparing models with ($M_{si}$) and without ($M_{ft}$) p\textsuperscript{2}-TQA.
We randomly sample 50 instances from HiTab and manually examine the correctness of reasoning steps for both Qwen and LLaMA, yielding a total of 200 reasoning chains (50 instances × 2 models × 2 variants) and 245 steps for $M_{ft}$ versus 239 steps for $M_{si}$. 
Models enhanced with p\textsuperscript{2}-TQA demonstrate higher step accuracy than those without (83\% vs.\ 75\%), with most improvements arising from reduced errors in planning and numerical reasoning.

\section{Conclusions}
We have introduced a self-improvement framework p\textsuperscript{2}-TQA that uses process-based preference learning. Our framework effectively improves the performance of TQA models by up to 5\%. 
The resulting models demonstrate competitive performance compared to state-of-the-art TQA systems, which depend on huge LLMs and tool usage. Yet, models enhanced with p\textsuperscript{2}-TQA require five times less inference time.

\section*{Limitations}
First, While our method effectively enhances the performance of small-sized TQA models, its impact on large TQA models remains unexplored. To the best of our knowledge, current fine-tuned TQA models only focus on small-sized LLMs. Future work can explore an efficient training strategy for large fine-tuned TQA models. 
Second, we limit the task in our study to only TQA, while there exist other table-related tasks, such as table summarization. Third, although our framework supports iterative self-learning, the present work only demonstrates the effectiveness of the first iteration, leaving multi-iteration evaluations for future study.
As the datasets we used in this study are originally sourced from Wikipedia, scientific papers, and statistical reports, we do not observe any potential risks from the datasets.



\bibliography{custom}
\bibliographystyle{acl_natbib}

\appendix

\section{Appendix}
\label{sec:appendix}

\subsection{Prompts}
\label{prompts}
\begin{figure*}
    \centering
    \includegraphics[width=1\linewidth]{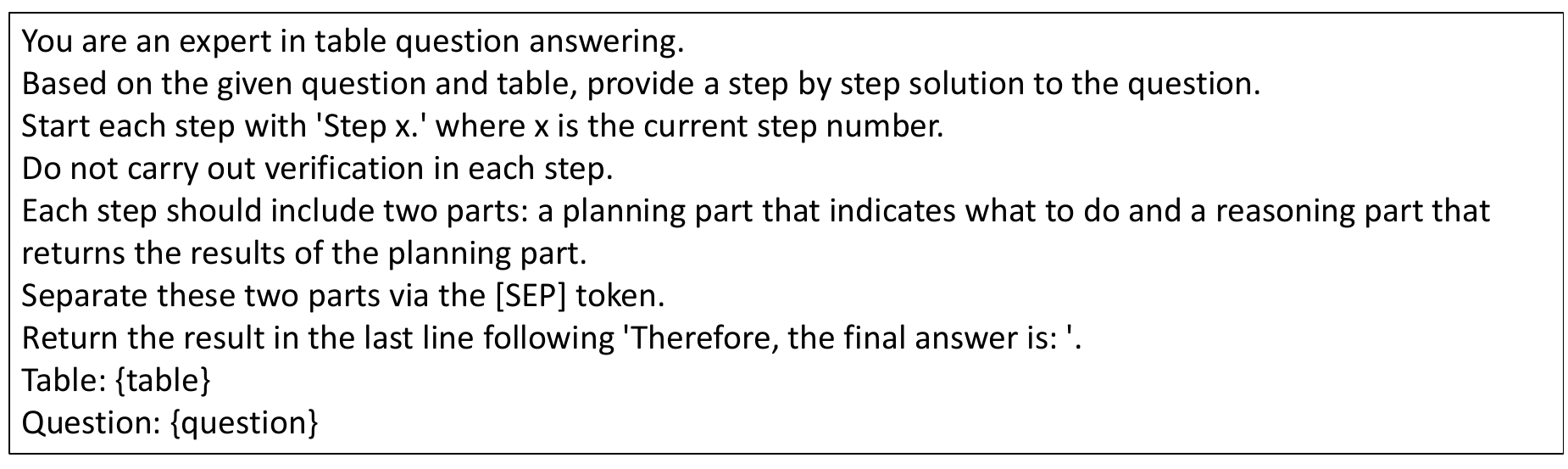}
    \caption{Prompt to generate full reasoning trace given a TQA problem.}
    \label{fig:p1}
\end{figure*}

\begin{figure*}
    \centering
    \includegraphics[width=1\linewidth]{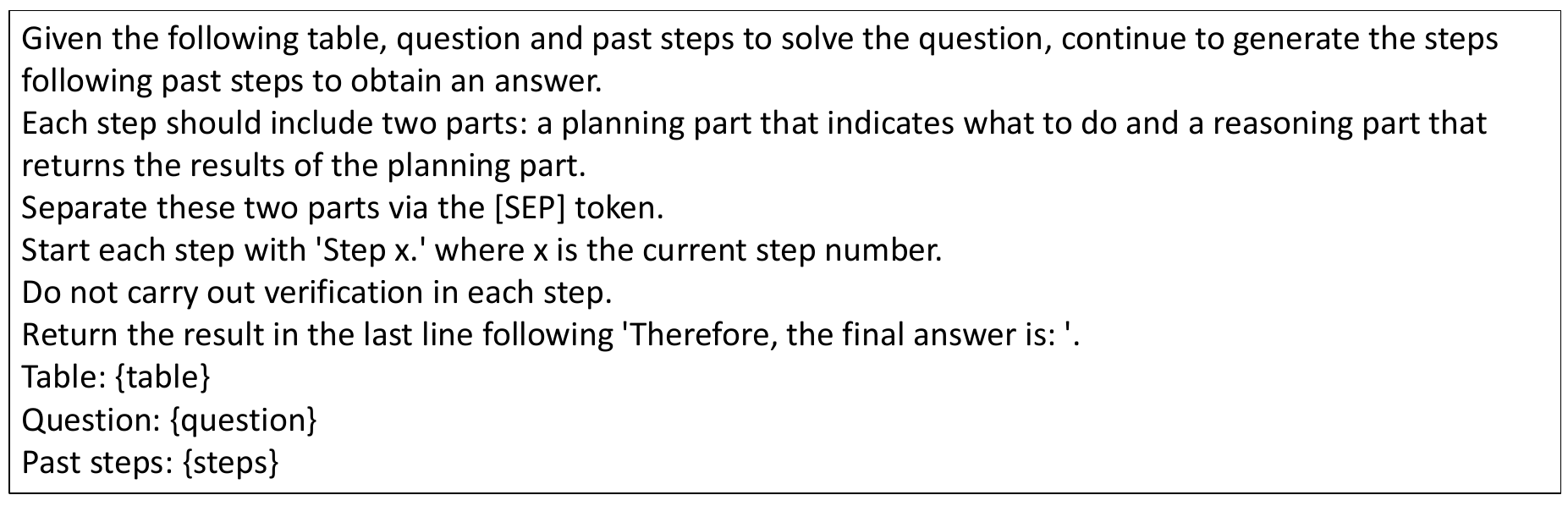}
    \caption{Prompt to complete a reasoning trace given a TQA problem and past steps.}
    \label{fig:p2}
\end{figure*}

\begin{figure*}
    \centering
    \includegraphics[width=1\linewidth]{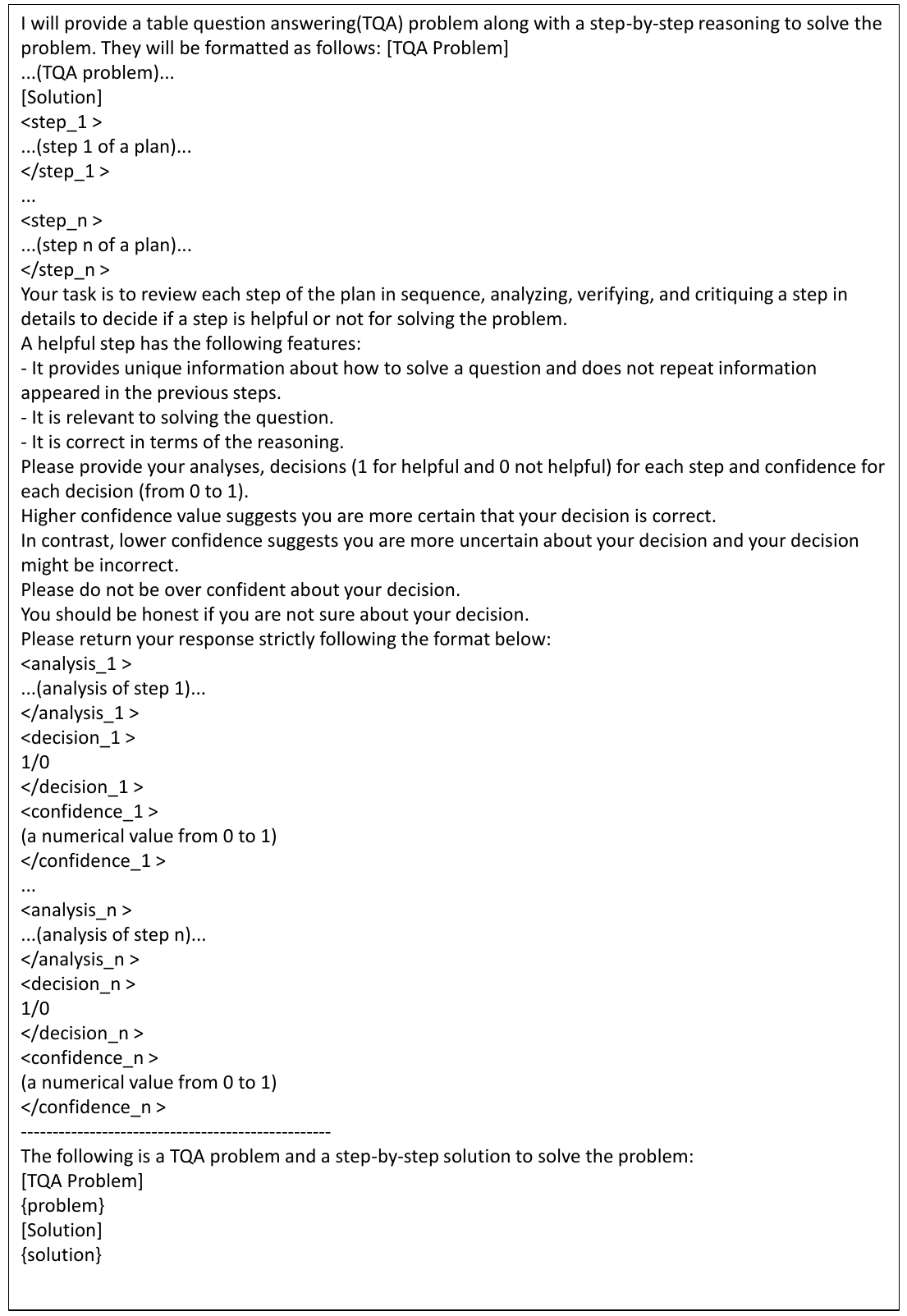}
    \caption{LLM judge prompt to generate analysis, decisions and confidence for each step.}
    \label{fig:p3}
\end{figure*}

Figure \ref{fig:p1}, \ref{fig:p2}, and \ref{fig:p3} show prompts for generating a full reasoning trace, completing a reasoning trace, and LLM judge evaluation for a reasoning trace.



\subsection{Datasets}
\label{datasets}
Table \ref{tab:test_statistics} shows the number of instances and domains for the test data we used.
WTQ \cite{pasupat-liang-2015-compositional}, HiTab \cite{cheng-etal-2022-hitab} and WikiSQL \cite{zhongSeq2SQL2017} are under the license of \textsc{CC-BY-SA-4.0}\footnote{\url{https://creativecommons.org/licenses/by-sa/4.0/}}, \textsc{BSD-3 Clause}\footnote{\url{https://opensource.org/license/bsd-3-clause}} and \textsc{C-UDA}\footnote{\url{https://github.com/microsoft/HiTab?tab=License-1-ov-file}} respectively.
TabFact \cite{2019TabFactA}, CRT \cite{zhang-etal-2023-crt} and SCITAB \cite{lu-etal-2023-scitab} are under the \textsc{MIT}\footnote{\url{https://opensource.org/license/mit}} license.

\begin{table}[!h]
    \setlength{\tabcolsep}{6pt}
    \centering    
 
    \begin{tabular}{ccc}
    \toprule
    Datasets & \#instances & Domain \\
    \midrule
    WTQ & 4344& Wikipedia \\
    TabFact & 12779 & Wikipedia\\
    HiTab & 1584 &statistical reports\\
    WikiSQL & 15878 &Wikipedia \\
    SCITAB &1224 &  scientific paper \\
     CRT & 728 &Wikipedia  \\

    \bottomrule
    \end{tabular}
       \caption{Test data statistics. The second column shows the number of test instances in each dataset.}
       \label{tab:test_statistics}
\end{table}

\subsection{Sampled Dataset Statistics}
\label{sampled_data}
Table \ref{tab:dataset_statistics} shows the sampling size for each method. We find \textsc{mc-b} results in the most data while RFT the least.

     

\begin{table}[!h]
    \centering    
    \setlength{\tabcolsep}{2pt}
    \small
    \begin{tabular}{lcccc}
    \toprule
    Methods & HiTab & WTQ& TabFact& Total \\ \midrule
    Original& 1.6k&2.8k&5k&9.4k \\ \midrule
    RFT &2.2k/1.6k&2.7k/2.4k&3k/2.7k&8k/6.7k  \\
    FDPO &3.6k/2k &4.7k/4k&4.6k/3.6K&12.9k/9.6k  \\
    SDPO \\
    \hspace{2mm} +\textsc{mc-b} &25k/20.8k&46k/47.1k&37k/38.4k&109k/106k \\
    \hspace{2mm} +\textsc{mix} & 5.4k/3.5k&9.6k/9.7k&5k/4.9k&20k/18.1k \\
    \hspace{2mm} +\textsc{mc-0.9} & 6.8k/1.6k&16k/4.4k&11k/2.4k&33.8k/8.4k \\

    \bottomrule
    \end{tabular}
       \caption{Sampled dataset sizes for different methods. Results for Qwen-2.5-7B and LlaMA-3.1-8B are separated by ``/''. \textsc{mc-b} refers to using Monte Carlo sampling with binary values as the value function. \textsc{mix} stands for using both \textsc{mc-b} and an LLM judge (Qwen-2.5-72B) as the value function. \textsc{mc}-0.9 stands for using Monte Carlo sampling with continuous values and setting the selection threshold as 0.9. RFT refers to rejected sampling fine-tuning. FDPO and SDPO stand for full-trace DPO and step-wise DPO. \label{tab:dataset_statistics}}
\end{table}

\subsection{Hyper-parameters}
\label{hyperparameters}
Table \ref{tab:hyperparameters} shows the hyper-parameters used for model fine-tuning.
\begin{table*}[t]
    \small
    \setlength{\tabcolsep}{4pt}
    \centering    
 
    \begin{tabular}{lccccccc}
    \toprule
    Models & Method & Fine-tuning & Learning rate& Epoch & Batch size & LoRA rank & DPO $\beta$ \\
    \midrule
    Qwen-2.5-7B & supervised fine-tuning& full-parameter &  5e-6 & 2 & 128 & - & - \\
    Qwen-2.5-7B & rejected sampling fine-tuning & LoRA & 1e-5 & 1 & 128 & 64 & - \\
    Qwen-2.5-7B & full chain DPO & LoRA &  1e-5 & 3 & 128 & 64 & 0.1 \\
    Qwen-2.5-7B & step-wise DPO & LoRA & 1e-5 & 3 & 128 & 64 & 0.1 \\
    \midrule
    LlaMA-3.1-8B &supervised fine-tuning &  full-parameter &  5e-6 & 2 & 128 & - & - \\
     LlaMA-3.1-8B & rejected sampling fine-tuning &LoRA  &  1e-5 & 1 & 128 & 32 & - \\
     LlaMA-3.1-8B & full chain DPO &LoRA &  1e-5 & 3 & 128 & 32 & 0.1 \\
     LlaMA-3.1-8B & step-wise DPO &  LoRA & 1e-5 & 3 & 128 & 32 & 0.1 \\
     
    \bottomrule
    \end{tabular}
       \caption{Hyper-parameters used for model fine-tuning.}
       \label{tab:hyperparameters}
\end{table*}


\subsection{Additional Results}
\label{additional_results}
Table \ref{tab: results} shows different models' performance on the six investigated datasets under greedy decoding. 
To validate the effectiveness of applying p\textsuperscript{2}-TQA, we set the temperature to 0.8 during inference. We report the mean and standard deviation over five runs in Table \ref{tab:sampled_results}.

\begin{table*}[t]
    \small
    \setlength{\tabcolsep}{4pt}
    \centering    
 
    \begin{tabular}{lcccccccc}
    \toprule
    Models & WTQ & TabFact& HiTab & WiKiSQL &SCITAB & CRT & In-domain & Out-of-domain \\
    \midrule
    Qwen-2.5-7B & 28.66 & 73.77&26.07&47.76&39.46&35.58&42.83&40.93\\
    \hspace{2mm} + TQA training ($M_{ft}$)&54.93 & 82.37& 61.36 & 68.26 & 54.90 &48.49&66.22&57.22\\
    \hspace{4mm} +RFT&56.26 & 82.44 &60.29 &68.15&52.29&46.15&66.33&55.53 \\
      \hspace{4mm} +FDPO &61.10 & 84.65 &63.83& \textbf{72.04}  &\textbf{56.94}&\underline{52.06}&69.86&\textbf{60.34}\\
     \hspace{4mm} +\textsc{Self-Explore} &52.70 & 82.30 &55.62&  65.20  &53.67&44.23 &63.54 &54.37 \\
    \hspace{4mm} \textsc{mc-b} & 60.80 & 82.98 & \underline{65.72} & 70.08&52.04&50.24&69.83&57.45\\
    \hspace{4mm} \textsc{mix}& \textbf{63.86} & \textbf{85.32} & 64.20 & 71.44&\underline{55.56}&\textbf{52.47}&\underline{71.63}&59.82\\
     \hspace{4mm} \textsc{p\textsuperscript{2}-TQA} ($\tau=0.9$) & 63.10 & \underline{84.88} & \textbf{67.55} & \underline{71.97}&\textbf{56.94}&51.37&\textbf{71.84}&\underline{60.09}\\
     \midrule

    LlaMA-3.1-8B & 30.64 & 63.91&26.20&31.87&43.38&32.55&40.25&35.93\\
    \hspace{2mm} + TQA training ($M_{ft}$)&64.80 & 83.36& 69.63 & 69.48 & 51.63 &49.04&72.60&52.76\\
    \hspace{4mm} +RFT&62.06 & 84.15 &67.30 &\underline{70.73}&50.49&48.63&70.84&56.62 \\
      \hspace{4mm} +FDPO &65.22 & \textbf{85.91} &68.18& \textbf{71.34}  &\underline{53.35}&\underline{50.41}&73.10&\textbf{58.37}\\
     \hspace{4mm} +\textsc{Self-Explore} &61.67 & 82.83 &64.52&  67.79  &52.20&44.28 &69.67 &54.76 \\
         
    \hspace{4mm} \textsc{mc-b} & \underline{65.54} & 84.01 & \underline{69.95} & 70.67&50.25&49.31&\underline{73.17}&56.74\\
    \hspace{4mm} \textsc{mix}& 62.39 & 84.98 & 66.79 & 68.31&52.94&\textbf{51.24}&72.05&57.50\\
     \hspace{4mm} \textsc{p\textsuperscript{2}-TQA} ($\tau=0.9$) & \textbf{65.84} & \underline{85.50} & \textbf{70.33} & 70.08&\textbf{54.44}&50.27&\textbf{73.98}&\underline{58.26}\\

 
    \bottomrule
    \end{tabular}
       \caption{Exact Match accuracies of models fine-tuned with different strategies and value functions, generated with greedy decoding. We bold the best results and underline the second best results for each model type.\label{tab: results}}
\end{table*}

\begin{table*}[t]
    \small
    \setlength{\tabcolsep}{4pt}
    \centering    
 
    \begin{tabular}{lcccccc}
    \toprule
    Models & WTQ & TabFact& HiTab & WiKiSQL &SCITAB & CRT  \\
    \midrule
    Qwen-2.5-7B ($M_{ft}$) & 57.00 $\pm{0.46}$
 & 81.85 $\pm{0.22}$
&61.76 $\pm{0.22}$
&67.68 $\pm{0.22}$
&52.16 $\pm{1.13}$
&46.92 $\pm{1.15}$\\
\hspace{4mm} +RFT&51.08 $\pm{0.60}$
 & 80.28 $\pm{0.32}$
 &56.92 $\pm{0.55}$
&61.77 $\pm{0.08}$
&49.90 $\pm{0.83}$
&46.02 $\pm{0.60}$\\
\hspace{4mm} +FDPO &58.54 $\pm{0.23}$
 & 84.32 $\pm{0.28}$
 &61.40 $\pm{0.49}$
& 68.20 $\pm{0.23}$
& \textbf{56.98} $\pm{0.80}$
&49.64 $\pm{1.40}$\\

\hspace{4mm} \textsc{mc-b} & 57.36 $\pm{0.33}$
 & 82.74 $\pm{0.04}$
 & \underline{62.63} $\pm{0.65}$
 & 66.39 $\pm{0.17}$
&52.71 $\pm{0.71}$
&48.71 $\pm{1.00}$\\
\hspace{4mm} \textsc{mix}& \textbf{60.43} $\pm{0.57}$
 & \underline{84.79} $\pm{0.10}$
 & 62.58 $\pm{0.41}$
 & \underline{68.59} $\pm{0.14}$
&51.98 $\pm{1.05}$&\textbf{52.09} $\pm{0.37}$\\
\hspace{4mm} \textsc{p\textsuperscript{2}-TQA} & \underline{60.42} $\pm{0.29}$
 & \textbf{84.81} $\pm{0.21}$
 & \textbf{65.05} $\pm{0.80}$
 & \textbf{68.90} $\pm{0.13}$
& \underline{55.08} $\pm{1.04}$ &\underline{50.99} $\pm{0.80}$\\
\midrule

LlaMA-3.1-8B ($M_{ft}$) &57.67 $\pm{0.38}$  & 82.82 $\pm{0.29}$& 63.18 $\pm{0.80}$& 62.17 $\pm{0.18}$& 49.98 $\pm{1.34}$&47.86 $\pm{0.65}$\\
\hspace{4mm} +RFT&58.27 $\pm{0.47}$
 & 81.44 $\pm{0.15}$
 &63.14 $\pm{0.70}$
&65.47 $\pm{0.22}$
&47.83 $\pm{0.98}$
&47.91 $\pm{1.61}$\\
\hspace{4mm} +FDPO &\underline{62.18} $ \pm{0.41}$ & \underline{85.52} $\pm{0.34}$ &65.83 $\pm{0.43}$
& \textbf{68.12} $\pm{0.18}$
  & \underline{52.94} $\pm{0.39}$ & 
\textbf{50.41} $\pm{0.60}$
\\

\hspace{4mm} \textsc{mc-b} & 60.92 $\pm{0.18}$
 & 83.98 $\pm{0.17}$
 & \underline{65.86} $\pm{0.69}$
& \underline{65.86} $\pm{0.30}$
&52.25 $\pm{1.54}$
&49.07 $\pm{0.83}$\\
\hspace{4mm} \textsc{mix} & 59.57 $\pm{0.62}$
 & 84.99 $\pm{0.07}$
 & 62.11 $\pm{0.44}$
 & 63.23 $\pm{0.19}$ &51.42 $\pm{0.63}$
&50.05 $\pm{1.09}$
\\
\hspace{4mm} \textsc{p\textsuperscript{2}-TQA} & \textbf{62.66} $ \pm{0.55}$ & \textbf{85.71} $\pm{0.23}$ & \textbf{66.49} $\pm{0.79}$&65.83 $\pm{0.26}$ & \textbf{53.76} $\pm{0.56}$& \underline{49.15} $\pm{1.20}$\\

 
    \bottomrule
    \end{tabular}
       \caption{Exact Match accuracies of models fine-tuned with different strategies and value functions, generated with sampling. \textsc{p\textsuperscript{2}-TQA} significantly improve fine-tuned model. Compared to baselines, it achieves competitive  performance across in-domain datasets. \label{tab:sampled_results}}
\end{table*}

\begin{table}[t]
    \scriptsize
    \setlength{\tabcolsep}{3pt}
    \centering    
 
    \begin{tabular}{lccccc}
    \toprule
    Method & 2k & 4k &6k & 8k & 12k \\
    \midrule
 FDPO & 67.3$\pm{0.6}$ & 68.2$\pm{0.5}$&\textcolor{black}{68.5} $\pm{0.4}$&69.1$\pm{0.4}$  & \textcolor{black}{67.1}$\pm{0.3}$ \\
 p\textsuperscript{2}-TQA &\textcolor{black}{69.0} $\pm{0.5}$ & \textcolor{black}{69.7}$\pm{0.3}$&70.8$\pm{0.3}$& 71.4$\pm{0.2}$& 72.9$\pm{0.2}$\\

    \bottomrule
    \end{tabular}
       \caption{Exact Match against varying training sizes. Results are obtained by averaging across three runs and three in-domain datasets using Qwen-2.5-7B.}\label{tab:computing_budges}
\end{table}

\subsection{Cost Effectiveness Analysis}
\label{cost_effectiveness}
Note that step-wise sampling requires higher computing budgets than full chain sampling; we conduct a cost-effectiveness analysis over FDPO and p\textsuperscript{2}-TQA. 
We do that by examining models' performance under different computing budgets (approximately by training instances). 
We first calculate the average number of tokens needed to generate an instance for different methods. 
This results in approximately 3k for FDPO and 10K for p\textsuperscript{2}-TQA.
We fine-tune models with varying sizes of training samples.
This not only allows us to compare FDPO and p\textsuperscript{2}-TQA under the same computing budgets, but also demonstrates each method's sensitivity to training size.
Table \ref{tab:computing_budges} shows the results. 
We observe that for FDPO, adding more training data does not necessarily improving models' performance (69.1-->67.1).
In contrast, scaling training sizes remains effective for  p\textsuperscript{2}-TQA.
This suggests that though FDPO generates more training instances than our method under the same budgets, the real effect of the generated data on performance is limited.



\subsection{Reasoning Chains Analysis}
\label{judge_analysis}
\begin{figure*}
    \centering
    \includegraphics[width=0.9\linewidth]{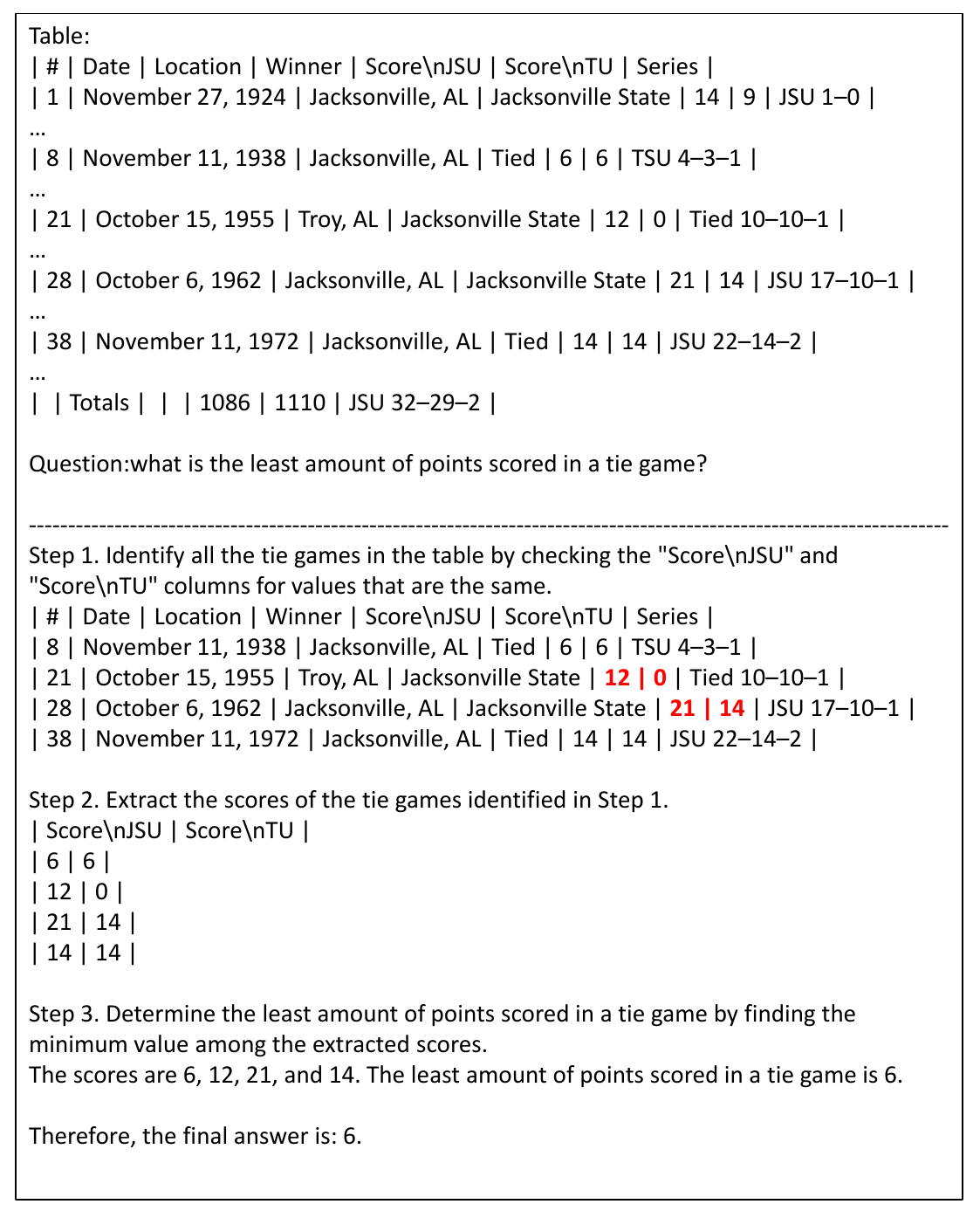}
    \caption{Wrong reasoning chain generated by a self-improved model (Qwen-2.5-7B) using \textsc{p\textsuperscript{2}-TQA}. The first wrong step is highlighted with red.}
    \label{fig:sc_error}
\end{figure*}

We sample 100 reasoning chains leading to correct answers generated from models using p\textsuperscript{2}-TQA and models using \textsc{mix} as the value function.  We manually examine the correctness of the reasoning chains. Among the 100 instances, we exclude 8 instances where either the answers are incorrect or the questions are ambiguous. We find similar accuracies of the reasoning chains generated from the aforementioned methods, with 95.7 and 94.6, respectively. This suggests the two methods do not differ much in terms of leading to correct reasoning chains. Nevertheless, wrong reasoning chains leading to correct answers still exist, possibly due to overly complex table inputs. An error case is shown in Figure \ref{fig:sc_error}.

\begin{figure}[htbp]
  \begin{tikzpicture}
    \begin{axis}[
      title={Thresholds Comparison with Soft Estimation},
      xlabel={Datasets},
      ylabel={EM},
      xmin=0, xmax=5,
      xtick={0,1,2,3,4,5},
      xticklabels={WTQ, TabFact, HiTab, WikiSQL, SCITAB, CRT},
      xticklabel style={rotate=60, anchor=north east},
      ymin=50, ymax=85,
      legend pos=north east,
      legend style={nodes={scale=0.8, transform shape}},
      width=0.48\textwidth, 
      height=0.8\linewidth, 
    ]

      \addplot[color=red,mark=none,line width=0.35mm] coordinates {
        (0,61.56) (1,83.86) (2,65.34) (3,70.47) (4,55.56) (5,50.41)
      };
      \addlegendentry{Soft Estimation ($\tau$=0.5)};

      \addplot[color=blue,mark=none,line width=0.35mm] coordinates {
        (0,62.62) (1,84.92) (2,67.80) (3,71.98) (4,55.47) (5,50.55)
      };
       \addlegendentry{Soft Estimation ($\tau$=0.7)};

      \addplot[color=green,mark=none,line width=0.35mm] coordinates {
        (0,63.10) (1,84.88) (2,67.55) (3,71.97) (4,56.94) (5,51.37)
      };
       \addlegendentry{Soft Estimation ($\tau$=0.9)};
    \end{axis}



  \end{tikzpicture}
  \caption{Thresholds comparisons with different value functions on six TQA datasets.}
  \label{fig:thresholds_compare}
\end{figure}

\subsection{Threshold Analysis}
\label{thresholds_analysis}
The threshold $\tau$ decides the state value differences when sampling a pair of (preferred and not preferred) states. We set $\tau$ to 0.9 in our study. We experiment with different values of $\tau$ to investigate its impact on the fine-tuned process-supervised models. The experimental settings are the same as described in Section \ref{sec:experimental_settings} except that we change the values of $\tau$. 
Figure \ref{fig:thresholds_compare} shows the performances of models fine-tuned with data sampled using different $\tau$. We observe that there is a tendency for higher thresholds to lead to better performance. However, we do not observe big differences in terms of model performances when setting $\tau$ to 0.7 or 0.9. 


\end{document}